\documentclass[10pt, a4paper]{article}
\pdfoutput=1
\usepackage{lrec2022}
\usepackage{multibib}
\newcites{languageresource}{Language Resources}
\usepackage{graphicx}
\usepackage{tabularx}
\usepackage{soul}
\usepackage{titlesec}

\titleformat{\section}{\normalfont\large\bfseries\center}{\thesection.}{1em}{}
\titleformat{\subsection}{\normalfont\SmallTitleFont\bfseries\raggedright}{\thesubsection.}{1em}{}
\titleformat{\subsubsection}{\normalfont\normalsize\bfseries\raggedright}{\thesubsubsection.}{1em}{}
\renewcommand\thesection{\arabic{section}}
\renewcommand\thesubsection{\thesection.\arabic{subsection}}
\renewcommand\thesubsubsection{\thesubsection.\arabic{subsubsection}}

\usepackage{epstopdf}
\usepackage[utf8]{inputenc}
\usepackage{textcomp}
\usepackage{hyperref}
\usepackage{xstring}
\usepackage[nameinlink]{cleveref}
\usepackage{color}
\PassOptionsToPackage{hyphens}{url}\usepackage{hyperref}

\title{Use of Transformer-Based Models for Word-Level Transliteration of the Book of the Dean of Lismore}

\name{Edward Gow-Smith\textsuperscript{1}, Mark McConville\textsuperscript{2}, William Gillies\textsuperscript{3}, \\ {\bf \large Jade Scott\textsuperscript{2}, Roibeard Ó Maolalaigh\textsuperscript{2}}}

\address{\textsuperscript{1}University of Sheffield, \textsuperscript{2}University of Glasgow, \textsuperscript{3}University of Edinburgh\\
         egow-smith1@sheffield.ac.uk, Mark.McConville@glasgow.ac.uk\\
         }

\abstract{
The Book of the Dean of Lismore (BDL) is a 16th-century Scottish Gaelic manuscript written in a non-standard orthography. In this work, we outline the problem of transliterating the text of the BDL into a standardised orthography, and perform exploratory experiments using Transformer-based models for this task. In particular, we focus on the task of word-level transliteration, and achieve a character-level BLEU score of 54.15 with our best model, a BART architecture pre-trained on the text of Scottish Gaelic Wikipedia and then fine-tuned on around 2,000 word-level parallel examples. Our initial experiments give promising results, but we highlight the shortcomings of our model, and discuss directions for future work.
 \\ \newline }

\begin{document}

\maketitleabstract

\section{Introduction}
As a material object, the Book of the Dean of Lismore (henceforth BDL) is a manuscript consisting of 159 paper folios, thought to have been assembled between 1512 and 1526 in eastern Perthshire, primarily by James MacGregor (c.1480–1551), the vicar of Fortingall and titular Dean of St. Moluag's Cathedral on Lismore \cite[59–60]{companion}. It is believed to have been acquired by James MacPherson, the Ossian `translator', from a Portree blacksmith around 1760, and was handed over to the Highland Society of Scotland in 1803. It is now located in the National Library of Scotland (Adv.MS.72.1.37).

As an information object, the BDL is primarily an eclectic collection of traditional Gaelic poetry, including bardic, heroic and informal verse, by diverse authors, both professional and amateur, Scottish and Irish. Perhaps the most notable feature of the manuscript is that the Gaelic verse was not written in the traditional, morphophonemic Gaelic system of orthography but rather in a heterodox, semi-phonemic system based on the one used for writing Scots at that time. Consider, for example, the following two versions of the first line of p.128:
\begin{itemize}
  \item{Ne wlli in teak mir a hest a zramm a der a weit trane}
  \item{Ní bhfuil an t-éag mar a theist, a dhream adeir a bhith tréan}
\end{itemize}
The first version is essentially the one that appears in BDL itself, and the second is a reconstruction of how this would have been written in the traditional Gaelic orthography of the time. Note the seventh word \emph{hest}:\emph{theist}. The initial consonant in this word would have been pronounced as the voiceless glottal fricative [h] and this is clearly reflected in the Scots-based orthography. However, the reconstructed Gaelic \emph{th} includes a representation of the underlying morphophoneme T which is associated with (at least) two different phonemes – the fortis /t/ (written as \emph{t}) and the lenis /h/ (written as \emph{th}). The vowel in the final word \emph{trane}:\emph{tréan} is another example – the vowel here is the front mid [e:], represented in Scots orthography using the discontinuous digraph \emph{a\_e} and in Gaelic as the (non-discontinuous) digraph \emph{éa}.

Over the last 100 years, attempts have been made to \textbf{transcribe} some of the poems in BDL (i.e. decode the handwriting) and then to \textbf{transliterate} these into some version of traditional Gaelic orthography, e.g. \cite{quiggin,ross,gillies,meek}. However, until recently an internally consistent transcription and transliteration of the full manuscript had not been attempted. Since BDL is an indispensable part of the textual foundation for the \emph{Faclair na Gàidhlig} project, which aims to create a comprehensive dictionary of Scottish Gaelic on historical principles, this has now become a priority. This paper reports on the first two phases of this work: (a) the production of a consistent transcription of the full BDL; and (b) initial experiments in constructing an automatic transliterator from the Scots-based orthography into traditional Gaelic orthography using a small amount of parallel training data.

\section{Data}
The work on creating a consistent digital transcription of the whole of BDL was undertaken by the third and fourth listed authors. The first phase of this project involved digitally re-transcribing the manuscript transcription of BDL produced by Rev. Walter McLeod in 1893, when the BDL folios were in better physical condition than they are nowadays (NLS MS.72.3.12). Once this had been completed, a second iteration involved comparing this digital transcription with the handwriting in the BDL itself, in order to identify and correct any apparent errors in McLeod’s manuscript. (We are grateful to NLS for providing us with high-resolution digital images of both manuscripts.) In creating the digital transcription, a standard set of Unicode character points was used to encode non-ASCII glyphs in the BDL. In general, scribal contractions were not expanded. Some light markup was included for scribal insertions and deletions, and page and line numbers. 

In order to provide some training data for our automatic transliterator, the third listed author provided reconstructed `Dean's Text' transliterations for twelve of the poems in the BDL. Due to the small amount of data available, we decided to run experiments on word-level transliteration. Thus, the original transcriptions and reconstructed transliterations were aligned, where possible, at the word level. The majority of the data is word-to-word transliterated, but there are some cases where one word in the BDL is transliterated into multiple words in Scottish Gaelic, and vice versa, making up 7.4\% of the data. A discussion of the shortcomings of this approach is given in \Cref{subsec:whole-sequence-transliteration}. In total there were 1,962 examples, and 50 examples were randomly selected to give eval and test sets.

\section{Experiments}
We are interested in transliterating from the BDL to Scottish Gaelic (henceforth referred to as bdl-gd) and vice versa (likewise referred to as gd-bdl), although the first direction is of greater practical importance. Character-level BLEU score \cite{papineni2002bleu} is used as an evaluation metric. We ran experiments on this task using Transformer-based models, implemented in Fairseq \cite{ott-etal-2019-fairseq}\footnote{We release our data and scripts for running our experiments at \url{https://github.com/edwardgowsmith/transliteration-book-of-the-dean-of-lismore}.}. For all experiments, tokenisation was performed at the character-level. The maximum sequence length was set at 20, to cover all of the available data whilst keeping computational requirements low. We also set the batch size at 1 due to the limited size of the training data, and the known problem of poor generalisation with large batch sizes \cite{keskar2016large}. For all of our models, the best performing model (by epoch) on the eval set was taken and evaluated on the test set. Full results are shown in \Cref{table:bleu-results}, and in the rest of this section we discuss the various models and approaches used.

\begin{table*}[ht]
\footnotesize
\def\arraystretch{1.1}
\centering
\begin{tabular}{| c | c | c | c | c |}
\hline
 & \multicolumn{2}{c|}{bdl-gd} & \multicolumn{2}{c|}{gd-bdl} \\
\cline{2-5}
Model & eval & test & eval & test \\
\hline
Transformer (tiny) & 35.32 & 41.16 & 30.17 & \textbf{46.26} \\
BART (tiny) & 44.93 & 38.64 & 21.04 & 22.18 \\
BART (base) & 58.68 & 53.32 & 36.17 & 30.15 \\
BART (base) + p/t longer & \textbf{62.47} & 53.75 & 36.77 & 38.88 \\
BART (base) + p/t longer + f/t longer & 59.46 & 52.09 & \textbf{36.94} & 34.68 \\
BART (base) + p/t longer + homophones & 59.60 & \textbf{54.15} & 34.75 & 31.77 \\
\hline
\end{tabular}
\caption{\label{table:bleu-results} Character-level BLEU scores of the models on the eval and test splits. Best results are shown in bold.}
\end{table*}

\subsection{Parallel Data Only}
Our first experiments were using just the available parallel data. We trained a Transformer \cite{vaswani2017attention} architecture with 2 layers and 2 attention heads for the encoder and decoder, and an embed dimension of 64, referred to as Transformer (tiny). We experimented with larger architectures, but found they were unable to learn from the available data. Our model was trained for 100,000 updates ($\sim$52 epochs), with a linear warm-up of the learning rate for 4,000 updates to 5e-4, then a linear decay to zero. We used the Adam optimizer \cite{kingma2014adam} with $\epsilon=$1e-6, $\beta=(0.9,0.98)$. On bdl-gd, this model achieved BLEU scores of 35.32 on the eval set and 41.16 on the test set. On gd-bdl, this model achieved BLEU scores of 30.17 on the eval set and 46.26 on the test set (\Cref{table:bleu-results}).

\subsection{Monolingual Pre-Training}
\label{subsec:monolingual-pretraining}
The next approach was to utilise monolingual Scottish Gaelic data for the task, so that the model would hopefully learn something of Scottish Gaelic orthography. For this, we used the text of Scottish Gaelic Wikipedia\footnote{\url{https://gd.wikipedia.org/}}, split to the word level, giving $\sim$600,000 words. We then pretrained BART \cite{lewis2019bart} architectures with the denoising task on this data. We first implemented a model with 2 layers, 2 attention heads, and embed dimension of 64 (referred to as BART (tiny) in reference to the Transformer model). We trained this model for 100,000 updates ($\sim$43 epochs). This model was then fine-tuned on the parallel training data, with the same hyperparameters as for Transformer (tiny). On bdl-gd, this model achieved BLEU score of 44.93 on the eval set, performing better than Transformer (tiny), and 38.64 on the test set, performing worse than Transformer (tiny). On gd-bdl, this model achieved BLEU scores of 21.04 on the eval set and 22.18 on the test set (\Cref{table:bleu-results}), performing significantly worse than Transformer (tiny). It is expected that pre-training on monolingual Scottish Gaelic data will not be of help in this direction, but the significantly worse performance is surprising (see \Cref{sec:discussion}).

We next tried the default BART (base) architecture, consisting of 6 layers, 12 attention heads, and an embed dimension of 768. On bdl-gd, this model achieved BLEU scores of 58.68 on the eval set and 53.32 on the test set, significantly outperforming Transformer (tiny). On gd-bdl, this model achieved BLEU scores of 36.17 on the eval set and 30.15 on the test set.
We also ran the same model with additional pretraining, up to 400,000 updates ($\sim$172 epochs), which has been shown to be of benefit to other Transformer-based models \cite{liu2019roberta}. On bdl-gd, this model achieved BLEU scores of 62.47 on the eval set and 53.75 on the test set, showing an increase in performance on both. On gd-bdl, this model achieved BLEU scores of 36.77 on the eval set and 38.88 on the test set, also showing an increase in performance on both (\Cref{table:bleu-results}). We also experimented with finetuning for longer (also 400,000 updates compared to 100,000), but this was found to lead to a general decrease in performance in both directions, although it did improve the performance on the eval set for gd-bdl (\Cref{table:bleu-results}).

\subsection{Data Augmentation}
Next, approaches were taken at augmenting the available training data, a common approach in low-resource neural machine translation \cite{haddow2021survey}. Since we are interested in word-level transliteration, and thus a word may be transliterated into a homophone of the provided example with a different spelling (specifically, a heterograph), we took an approach to augment the training data with homophones. We used IPA information for Scottish Gaelic provided by English Wiktionary\footnote{\url{https://en.wiktionary.org/}} - the data was parsed in order to find homophones for words it the training data. Unfortunately, IPA information was only available for a small number of items, which increased the training data from 1,862 to 1,938 examples (an increase of $\sim$4\%). With the addition of this augmented training data, the BLEU score of BART (base) on the eval set decreased (from 62.47 to 59.60), but the BLEU score on the test set increased (from 53.75 to 54.15), which makes sense as the introduction of heterographs should allow the model to generalise better (although we note that the increase in performance is small). Interestingly, this model performs significantly worse in the reverse direction, with BLEU scores of 34.75 and 31.77 on the eval and test sets, respectively (discussed in \Cref{sec:discussion}). It should be noted that this approach assumes that heterographs in modern Scottish Gaelic were also heterographs at the time of the BDL, which should be a valid assumption. An alternative approach to augmenting the data would be to use a rule-based approach, which we leave to future work.

\section{Discussion}
\label{sec:discussion}
In this section we discuss our results. From \Cref{table:bleu-results} we can see that, in general, the performance on gd-bdl is significantly worse than that on bdl-gd. This is to be expected, since the models have access to a large amount of monolingual Scottish Gaelic (gd) data, but BDL (bdl) is effectively an unseen language, which previous work has shown results in poor performance (see e.g. \newcite{ustun-etal-2021-multilingual}). What is perhaps unexpected, however, is that our best-performing model on bdl-gd, BART (base) + p/t longer + homophones, performs significantly worse than the best in the opposite direction (31.77 compared to 46.26 on the test set). In fact, our best-performing model on gd-bdl, Transformer (tiny), does not use any monolingual Scottish Gaelic data. It seems likely that our models are overfitting on the train and eval sets, as a result of their small sizes. Attempts to avoid this could be made, including using multi-fold cross-validation. Additionally, it is hoped that we will have access to more parallel data in the future which will alleviate this problem, as well as the variance of performance across the eval and test splits.

\subsection{Error Analysis}
In this section, we perform an error analysis by taking our best-performing model and investigating which examples in the test set this model performed worse on (by character-level BLEU score). These are shown in \Cref{table:error-analysis}. We note that these examples are relatively long; for shorter examples, our model generally performs better, which is typically expected but likely exaggerated in this case due to the increasing ambiguity of a word in the BDL as length increases. We note that our model struggles with spaces: no space is added when transliterating ``eflay'', and a space is erroneously added when transliterating ``waiwill'' (although the space is correctly removed when transliterating ``dwgis i''). Since examples containing spaces on either the source or target side only make up a small amount of the parallel data, and the pretraining data contains no spaces, this is an expected area of difficulty, which we discuss further in \Cref{subsec:handling-spaces}. We also note that, out of the seven examples here, our model appears to output only three true Scottish Gaelic words (``mha fháil'' meaning ``if found'', ``chuaiseach'' meaning ``cavities'', and ``mhíos'' meaning ``month''). This is not necessarily a problem, since we want our model to be able to output unseen words, for example old-fashioned spellings and proper nouns. However, contextual information may help to determine the validity of a given transliteration, though the limited data available may prove to limit the efficacy of such an approach. Interestingly, the model transliterates ``di'' as the ``[UNK]'' token, which is problematic.

\subsection{Learning of Scottish Gaelic Spelling Rules}
We note that all of the outputs from our best model are \emph{plausible} words, in that they obey the spelling rules of Scottish Gaelic. This is not the case for the Transformer (tiny) model trained only on the parallel data --- as an example ``dwgis'' is transliterated by this model into ``duigas'', which is not an acceptable Scottish Gaelic word, since a medial consonant must be surrounded by vowels of the same type \cite{gillies2009scottish}. This suggests that the training on monolingual data has allowed our model to learn the rules of Scottish Gaelic spelling, which has in turn improved performance on the transliteration task.

\begin{table}[ht]
\footnotesize
\def\arraystretch{1.1}
\centering
\begin{tabular}{| c | c | c |}
\hline
Input & Output & Reference \\
\hline
eflay & e’léamh & a’ phláigh \\ 
dwgis i & duise & dtugas-sa \\ 
chotly\textsuperscript{t}sy\textsuperscript{t} & chuaiseach & chodlas-sa \\
wawaill & mha fháil & bhfaghbha'il \\ 
deina\=r & díonar & d’éinfhear \\ 
fean\=e & fén & phéin \\ 
zonicht & dhuanancht & dhona \\
di & [UNK] & do \\
gawe & gáimh & gabh \\
weiß\textsuperscript{t} & mhíos & bhíos \\ 
\hline
\end{tabular}
\caption{\label{table:error-analysis} The ten examples that our best performing model performed worse on for the test split (from bdl-gd).}
\end{table}

\section{Future Directions}
Our preliminary experiments have shown promise in the task of transliterating the BDL, however there are many areas for improvement that we hope to address in future work. 

\subsection{Whole Sequence Transliteration}
\label{subsec:whole-sequence-transliteration}
Since our work here is on word-level transliteration, it is unclear how this will extend to longer sequences, especially in the case of many-to-one transliteration. We take an example of transliterating a whole sequence with our model, shown in \Cref{table:whole-sequence}.

\begin{table}[ht]
\footnotesize
\def\arraystretch{1.1}
\centering
\begin{tabular}{| c | c | c |}
\hline
Input & A wēni\textsuperscript{t} za dwgis i grawġ \\
Output & a bhean dhá duis a’ grádh\\
Reference & A bhean dhá dtugas-sa grádh \\        

\hline
\end{tabular}
\caption{\label{table:whole-sequence} Transliterating a whole sequence with our model.}
\end{table}

In order to transliterate this whole sequence, we split it on whitespace and then pass each word individually to the model. Since, in this case, ``dwgis i'' is transliterated into a single word, our model cannot capture this (although note that this model fails to correctly transliterate these two words anyway (see \Cref{table:error-analysis})). An alternative approach to transliterating multi-word sequences may therefore be needed. Currently, due to our models being set at a max sequence length of 20, longer sequences cannot be directly given to the model.

\subsection{Handling of Spaces}
\label{subsec:handling-spaces}
A related problem is the tendency of the models to struggle with handling spaces, both in the case of one-to-many and many-to-one transliteration. In order to help with this problem, it is likely we will need to include examples containing spaces during pre-training, or perform oversampling on the available training data to balance the number of examples with spaces and those without. 

\subsection{Data for Pre-Training}
As stated in \Cref{subsec:monolingual-pretraining}, we used data from Scottish Gaelic Wikipedia for pretraining, which is written in standardised modern Scottish Gaelic. For the purposes of our task, we are interested in generating transliterations which are faithful to the pronunciation at the time of the BDL. Hence, other data sources may provide more relevance for pre-training, such as Corpas na Gàidhlig\footnote{\url{https://dasg.ac.uk/corpus}} which contains transcribed texts dating back to the 17th century, and this is a direction of future work.  

\section{Related Work}
There is no previous work, to the best of our knowledge, that uses Transformer-based models for tasks involving Scottish Gaelic. However, such approaches have been applied to other languages in the Celtic family: multilingual BERT \cite{devlin-etal-2019-bert} contains Irish, Welsh and Breton in its training data, and there is a monolingual BERT for Irish \cite{barry2021gabert} which was shown to outperform multilingual BERT on a dependency parsing test. There have been previous approaches at applying Transformer-based models to the task of word-level transliteration. \newcite{wu-etal-2021-applying} applied the vanilla Transformer to the NEWS 2015 shared task \cite{zhang-etal-2015-whitepaper}, outperforming previous models. \newcite{singh2021neural} also applied various sizes of Transformer architectures to the task of transliterating Hindi and Punjabi to English.

\section{Conclusion}
In this paper we discuss approaches to training Transformer-based models on the task of transliterating the Book of the Dean of Lismore (BDL) from its idiosyncratic orthography into a standardised Scottish Gaelic orthography. In particular, we outline our preliminary experiments training these models for word-level transliteration using both parallel word-level transliteration data for finetuning and monolingual Scottish Gaelic data for pretraining. Our best performing model was able to achieve a character-level BLEU score of 54.15 on the test set, showing significant promise, although there are many directions for improvement and future work, including extending this work to sequence-level (multi-word) transliteration. 

\section{Acknowledgements}
This work was supported by \emph{Faclair na Gàidhlig} and the Centre for Doctoral Training in Speech and Language Technologies (SLT) and their Applications funded by UK Research and Innovation [grant number EP/S023062/1].

\section{Bibliographical References}\label{reference}

\bibliographystyle{lrec2022-bib}
\bibliography{lrec2022-example}

\end{document}